\documentclass[fleqn,10pt,twocolumn]{SWARM25}
\usepackage{graphicx} 
\usepackage{enumitem}
\usepackage{xcolor}
\usepackage{soul}
\usepackage{url}

\usepackage{amsmath,amssymb}

\newcommand{\celeg}{{\it C. elegans}}
\title{From Pheromones to Policies: Reinforcement Learning for Engineered Biological Swarms}

\author{Aymeric Vellinger${}^{1\dagger}$, Nemanja Antonic$^1$ and Elio Tuci$^1$}
\speaker{Aymeric Vellinger}

\affils{${}^{1}$Department of Computer Science, University of Namur, Namur, Belgium\\
(E-mail: aymeric.vellinger@unamur.be)\\
}
\abstract{ 
Swarm intelligence emerges from decentralised interactions among simple agents, enabling collective problem-solving. This study establishes a theoretical equivalence between pheromone-mediated aggregation in \celeg\ and reinforcement learning (RL), demonstrating how stigmergic signals function as distributed reward mechanisms. We model engineered nematode swarms performing foraging tasks, showing that pheromone dynamics mathematically mirror cross-learning updates, a fundamental RL algorithm. Experimental validation with data from literature confirms that our model accurately replicates empirical \celeg\ foraging patterns under static conditions. In dynamic environments, persistent pheromone trails create positive feedback loops that hinder adaptation by locking swarms into obsolete choices. Through computational experiments in multi-armed bandit scenarios, we reveal that introducing a minority of exploratory agents insensitive to pheromones restores collective plasticity, enabling rapid task switching. This behavioural heterogeneity balances exploration-exploitation trade-offs, implementing swarm-level extinction of outdated strategies. Our results demonstrate that stigmergic systems inherently encode distributed RL processes, where environmental signals act as external memory for collective credit assignment. By bridging synthetic biology with swarm robotics, this work advances programmable living systems capable of resilient decision-making in volatile environments.
 }


\begin{document}

\maketitle
\section{Introduction}

Collective animal behaviour illustrates how large groups accomplish sensing, decision making and actuation that far exceed the cognitive or energetic capacity of the individuals that compose them.  Flocks fuse noisy viewpoints into collision-free trajectories, termite mounds regulate climate without thermostats, and pheromone-based ant foraging solves dynamic optimisation on noisy terrain. These examples have inspired swarm intelligence, a biologically inspired, self-organised and decentralised problem-solving paradigm in which simple artificial agents encode local interaction rules abstracted from biological swarms, enabling the collective to exhibit coherent, adaptive and often near-optimal behaviour without any agent possessing global knowledge or acting as a central controller. This emergent collective cognition, defined as the suite of processes by which an entity acquires, transforms and utilises information to guide behaviour, would allows the swarm to process dispersed information to perform tasks that surpass the capabilities of its individual members.  

In artificial intelligence a parallel idea appears at the level of a single agent. Reinforcement learning (RL), as conceptualised by Sutton and Barto~\cite{sutton1998reinforcement}, synthesises psychological principles of Thorndike’s law of effect (trial-and-error learning), optimal control theory from engineering, and operant conditioning paradigms. Thus, what swarms exhibit at the population level and what a single RL agent exhibits at the individual level are two scales of the same cognitive learning dynamic: decentralised, incremental hypothesis testing driven by feedback from the environment. A direct consequence of treating learning as continual model revision is adaptability (i.e., the capacity to maintain competent behaviour under changing conditions). In animal conditioning, extinction denotes the decay of a conditioned response when reinforcement ceases. By analogy, weight decay, entropy regularisation, or uncertainty-aware Bayesian updates act as gradual dampers in RL, and more explicit strategies, such as policy pruning or experience forgetting buffers, can forcibly erase outdated behavioural schemas to make room for new ones. Balancing knowledge consolidation and extinction~\cite{quirk2008neural}, therefore amounts to regulating exploration–exploitation~\cite{yogeswaran2012reinforcement}. Achieving that balance is central to designing swarm systems that remain resilient, sample-efficient, and cognitively coherent over long horizons.

Even if swarm robotics shows promising engineering solutions, it falls short in facing natural environments~\cite{dorigo2021swarm}. This gap motivates the exploration of biological animal robots \cite{BABOTS_WebPage}, where living organisms themselves constitute the agents.
 In this context, \textit{Caenorhabditis elegans} (\celeg) with its fully mapped neural connectome consisting of $302$ neurons~\cite{white1986structure} emerges as a favorable candidate. 
 This neural simplicity, combined with emerging techniques in neuro-synthetic biology \cite{rabinowitch2019synthetic}, opens opportunities for creating bioengineered \celeg~ as swarm agents. By genetically manipulating synaptic connectivity \cite{rabinowitch2024understanding} and constructing in vivo gene circuits \cite{kukhtar2023synthetic}, it may becomes feasible to systematically alter collective behaviours, thus transforming \celeg~ populations into a programmable biological swarm of robots.

\celeg~ occurs in diverse natural and laboratory strains that span a behavioural continuum. The canonical laboratory reference strain $N2$ has been demonstrated to be almost entirely \emph{solitary}, whereas strains carrying the low-activity allele of the neuropeptide-receptor gene $npr-1$ adopt a more \emph{social} collective behaviour, i.e. clustering in clumps at the food interface \cite{DEBONO1998}. Nonetheless, even these \emph{social} worms exhibit little beyond simple aggregation, far from the elaborate tasks of eusocial or subsocial insects. It is conceivable, however, that pheromone-based communication could be harnessed to engineer much more elaborate collective behaviour, allowing us to treat \celeg\ populations as programmable biological robots whose coordination is mediated by tunable chemical signals.

In this context, our study forges a formal bridge between such engineered stigmergy (i.e., coordination via environmentally deposited traces such as pheromones) and RL theory, thereby embedding stigmergic swarms within the wider framework of collective cognitive learning. We then examine how environmental volatility and population heterogeneity shapes swarm adaptability performance, using pheromone-guided foraging as a collective decision problem. We demonstrate:

\begin{enumerate}
\item the mathematical equivalence between pheromone-driven swarming dynamics and the cross-learning updates rule, with in-vivo experimental validations.
\item the trade-off between consensus duration and adaptability when environment shifts;
\item the crucial role of a minority of exploration-oriented individuals in sustaining plasticity and long-term resilience, enabling extinction at the swarm level. \end{enumerate}

In summary, our study introduces a novel biologically-grounded framework that links stigmergic swarming to reinforcement learning, demonstrating how targeted genetic engineering of pheromone signalling pathways in \celeg~ could enhance collective decision-making and adaptability, thereby advancing the design of programmable, resilient biological swarms capable of sophisticated, adaptive behaviours in dynamic environments.

\section{Preliminaries}
\subsection{C. elegans foraging patterns}
\label{foraging}
The foraging behaviour of \celeg\ can be modelled using empirical data from controlled experiments that quantify nematode aggregation across bacterial patches of varying density~\cite{madirolas2023caenorhabditis}. In these experiments, populations of five nematodes are repeatedly released (58 trials) from a central location on agar plates containing equidistant bacterial patches. The observed distribution dynamics are aggregated to derive a phenomenological model describing patch selection as a function of bacterial density and strain.

For a food patch $i$ with bacterial density $D_i$, the relative attractiveness $A(D_i)$ follows a sigmoidal relationship:
\begin{equation}
A(D_i) = \sqrt{H} \cdot \frac{1 + 4\left(\tfrac{D_i}{D_{\mathrm{attract}}}\right)^{k}}{H + 4\left(\tfrac{D_i}{D_{\mathrm{attract}}}\right)^{k}},
\end{equation}
where $H$ defines the dynamic range (ratio of maximum to minimum attractiveness), $k$ governs the sigmoid steepness, and $D_{\mathrm{attract}}$ is the density at which attractiveness reaches fivefold the baseline. These parameters are strain-specific, with $H = 51.5$, $k = 0.29$, and $D_{\mathrm{attract}} = 0.003$ identified for \textit{E.~coli} OP50 via Bayesian optimisation fitting~\cite{madirolas2023caenorhabditis}.

The proportion $P_i$ of nematodes aggregating at patch $i$ across $M$ patches follows a softmax distribution:
\begin{equation}
P_i = \frac{A_i}{\sum_{j=1}^M A_j}.
\end{equation}
This formulation approximates an ideal free distribution under negligible inter-worm interactions~\cite{HOUSTON1987301}, validated experimentally through high goodness-of-fit ($R^2 > 0.95$) across diverse patch configurations.

\subsection{Reinforcement Learning and Multi-armed bandit}
The multi-armed bandit problem formalises the exploration–exploitation trade-off faced by decision-makers. Consider $n$ arms with hidden reward distributions $\{r(a)\}$. A reinforcement-learning agent maintains a policy  $\pi(t)=\bigl(\pi_{1},\ldots,\pi_{n}\bigr)$ representing action-selection probabilities. Cross Learning \cite{sutton1998reinforcement} updates $\pi$ after choosing arm $k$ with reward $r_k$ as:

\begin{equation}
\label{eq:croos-learning}
\pi_a(t+1)=\pi_a(t)+\alpha\, r_k
\begin{cases}
1-\pi_a(t), & \text{if } a = k,\\[6pt]
-\pi_a(t), & \text{otherwise}.
\end{cases}
\end{equation}

Here $\alpha$ is the learning rate, regulating the impact of the reward on the policy.  
This update rule emerges naturally from swarm-intelligence models via replicator dynamics \cite{sandholm2010population}.  
When a population of agents imitates successful neighbours (the voter rule), their collective behaviour is equivalent to a single RL agent performing Cross Learning \cite{soma2024bridging}.  
The policy vector $\pi$ mirrors population proportions, with individual interactions implementing stochastic gradient ascent on expected rewards, leading to the \textit{Taylor Replicator Dynamic}, defined as:
\begin{equation}
\label{replicator}
\dot{\pi}_a = \pi_a \left(q^\pi_a - v^\pi\right),
\end{equation}
where $\dot{\pi}_a$ is the derivative of the $a$-th component of the population vector $\pi_a$, $q^\pi_a := \mathrm{E}$$[r_a]$ is the expected payoff of type $a$ against the current population, and $v^\pi := \sum_b \pi_b \mathrm{E}$$[r_b]$ is the current average payoff of the entire population.
We will later demonstrate that in biological terms, the agent’s policy $\pi(t)$ reflects the population distribution across food patches, where inter-individuals attraction drives policy updates akin to reward-based learning.
\label{mab-cl}

\section{Model}
We develop our model based on the empirical framework described in Section \ref{foraging}, with the aim of integrating pheromone-based communication into a swarm of \celeg.
Let the attractiveness function for the pheromone at each site $i$: $B_i = \tau_i(t).$
Where $\tau_i$ is the quantity of pheromone in site $i$. Thus, in our formulation, the attractiveness of a site due to pheromones is assumed to scale linearly with the quantity of pheromone deposited.

However, it is important to acknowledge that this linear relationship may not fully represent biological reality. In \celeg, the sensory neurons responsible for detecting ascaroside pheromones, for example, exhibit nonlinear responses, with sensory saturation occurring at both very low and very high pheromone concentrations. Furthermore, experimental evidence indicates that worm responses to pheromones are dynamic and can vary temporally, potentially shifting from attraction to aversion depending on the physiological state or environmental context \cite{dal2021inversion}.

Due to current limitations in experimental data regarding these complex sensory dynamics, we adopt a simplified assumption of unbounded linear attraction to pheromone concentration. This simplification, while sacrificing some biological realism, allows us to explore fundamental principles within an analytically tractable and broadly applicable information-theoretic framework. This approach also sets the stage for future extensions where more detailed sensory response models and context-dependent valence shifts can be systematically incorporated.

With the introduction of the pheromone attractiveness function, the probability of worms aggregating at the $i^{\mathrm{th}}$ food patch, in an environment with $M$ food patches, can now be expressed as:
\begin{equation}
    P_i(t) = \frac{B_i(t) \, A_i}{\sum_{j=1}^M B_j(t) \, A_j},
\end{equation}
where $B_i(t)$ denotes the attractiveness due to pheromones at site $i$ at time $t$, and $A_i$ represents the intrinsic bacterial attractiveness of the site, as seen in Section \ref{foraging}.

Within this model, we ensure that pheromone attractiveness remains non-negative and never diminishes the intrinsic appeal of a food patch. Hence, we impose the condition $B_i(t) \geq 1$ for all $i$. Consequently, all patches are initialised with a baseline pheromone attractiveness of $B_i(0) = 1$.

Our objective now shifts toward understanding the temporal dynamics of worm distribution across patches. Specifically, we seek to characterise how the decisions made by individual worms influence subsequent choices within the swarm, driving the collective dynamics through stigmergic reinforcement.
Let the evolution of pheromone quantity $\tau_i(t)$ at the $i^{\mathrm{th}}$ patch over time defined by the following dynamics:
\begin{equation}
     \tau_{i}(t+1) = \rho \,\tau_{i}(t) +  \Delta_{k,i}(t),
\end{equation}
where the pheromone increment $\Delta_{k,i}(t)$ depends on individual decisions:
\begin{equation}
    \Delta_{i}(t) = 
\begin{cases}
Q, & \text{if agent $ k $ commits toward patch $ i $},\\[6pt]
0, & \text{otherwise}.
\end{cases}
\end{equation}

Here, the parameter $0 \leq \rho \leq 1$ represents the pheromone evaporation rate, modeling the gradual decrease of pheromone concentration over time. The parameter $Q \geq 0$ denotes the amount of pheromone deposited by an agent upon selecting a patch. Intuitively, these equations capture how the attractiveness of each patch naturally decays due to evaporation but increases incrementally whenever an individual chooses that patch, thus reinforcing its attractiveness.

\section{Learning}
In this section, we formally establish how the probabilities evolve through time when agents (worms) sequentially select food patches and secrete pheromones, connecting stigmergic dynamics with reinforcement learning principles.
Let us describe how the probabilities evolve over time when worms act sequentially, with the goal of rewriting it in terms of probability:
\begin{equation}
    P_i(t+1)= \frac{B_i(t+1)A_i(t+1)}{\sum_{j=1}^M B_j(t+1)A_j(t+1)}
\end{equation}
First, let's see how the system evolves when a worm secreting pheromone makes a decision toward a food patch:
\begin{equation}
     B_i(t+1) = \tau_i(t+1) = \rho\,\tau_i(t) + Q,
     \qquad
     A_i(t+1)= A_i
\end{equation}
Therefore, the bacterial attractiveness of a food patch doesn't evolve through time, so $A_i(t+1)= A_i(t)$, but the attractiveness related to pheromones evolves according to its quantity, which increases when a worm chooses this patch.

By replacing $B_i(t+1)$ we get:
\begin{equation}
  P_i(t+1)
  \;=\;
  \frac{\bigl[\rho\,\tau_i(t) + Q\bigr]\,A_i}
       {\rho\,\sum_{j=1}^M \tau_j(t)\,A_j 
        \;+\; 
        Q\,A_i}.
\end{equation}

After some mathematical manipulation (see Appendix \ref{deriv1}), we obtain:
\begin{equation}
    P_i(t+1)= P_i(t)+ \frac{QA_i(1-P_i(t))}{\rho\,\sum_{j=1}^M \tau_j(t)\,A_j + Q\,A_i}
\end{equation}

Similarly, when looking at the evolution of the probability of patch that is not chosen, we have:
\begin{equation}
    \tau_i(t+1)=\rho \tau_i(t).
\end{equation}
And therefore:
\begin{equation}
\begin{split}
P_i(t+1)
  &= \frac{\rho\,\tau_i(t)\,A_i}
     {\rho\,\sum_{j=1}^M \tau_j(t)\,A_j + Q\,A_k} \\[6pt]
  &= P_i(t)
     - \frac{Q\,A_k}{\rho\,\sum_{j=1}^M \tau_j(t)\,A_j + Q\,A_k}\,P_i(t).
\end{split}
\end{equation}
The derivation is in the Appendix \ref{deriv2}.

Finally, if we regroup the chosen and unchosen cases, we obtain:
\begin{equation}
\begin{aligned}
P_i(t+1)
  &= P_i(t) 
     + QR_i(t)
       \begin{cases}
         1 - P_i(t), & \text{if site $i$}, \\[6pt]
         -P_i(t),     & \text{otherwise}.
       \end{cases}
\end{aligned}
\end{equation}
with
\begin{equation}
    R_i(t)=\frac{A_i}{\rho\sum_{j=1}^{n}\tau_j(t)A_j + Q A_i}
\end{equation}
This update rule is mathematically equivalent to cross-learning (Eq. \ref{eq:croos-learning}), where environmental stigmergy implicitly implements reinforcement signals. Thus, swarm aggregation becomes a distributed RL process. 

Notably, this mechanism shapes collective behaviour: partial movements toward beneficial patches leave pheromone traces that further attract subsequent worms. Such incremental reinforcement guides the entire swarm to converge on optimal sites. This operant-like process occurs without any single individual requiring memory of prior outcomes; the environment’s stigmergic cues effectively store and communicate the evolving “collective knowledge.”

A limitation of purely positive feedback is the risk of lock-in if conditions change abruptly due to the slow extinction of collectively acquired knowledge. High pheromone persistence may prevent exploring new options unless pheromone evaporation $\rho$ is sufficiently high. Nonetheless, this idealised case shows how unbounded positive stigmergy can transform simple aggregation into distributed learning, where the environment itself encodes and updates knowledge, which is an emergent form of collective cognition reminiscent of operant conditioning.

In our model, worms act sequentially, depositing pheromones and updating patch attractiveness immediately upon selecting a site. While this may appear to deviate from simultaneous or batch decision-making commonly assumed in some swarm models, the assumption is theoretically justified by the ergodicity of cross-learning dynamics. In ergodic systems, time averages over sequential stochastic updates converge to the expected behaviour of an equivalent ensemble.
This ensures that the aggregated behaviour of the swarm remains statistically equivalent to the one of a population learning in parallel, supporting the robustness of our formal equivalence with cross-learning. In the following, we will constantly use a batch size of $100$.

Furthermore, this formal equivalence demonstrates that positive stigmergic systems operate analogously to classical voter models. In fact, it has been shown that the expected evolution of cross-learning dynamics coincides exactly with that of the voter model through the replicator dynamic \cite{soma2024bridging}. Hence, one may interpret a positive stigmergic system as a time‐dependent voting process in which each agent’s decision leaves a persistent influence via pheromone deposition, that shapes the probability of future choices.

\subsection{Buffer Replay}
In practice, a RL agent updates its policy based solely on immediate experiences, typically discarding previous rewards. Therefore, we need to approximate the overall quantity of pheromone in the environment ($\rho \sum_{j=1}^{n}\tau_j(t)$) with a replay buffer. A replay buffer stores historical interactions with the environment as tuples $(s_t, a_t, r_t, s_{t+1})$. At each learning step, we draw all past experiences from the store. To mimic evaporation, we implemented a \textit{Memory size} parameter, formally setting the maximum size of the replay buffer.

\section{Results}
We present three types of experiments using the reinforcement learning model. The first experiment aims to validate our model, while the other two focus on the swarm’s response in dynamic environments.
\subsection{Environments}
\begin{figure}
    \centering
    \includegraphics[width=0.9\linewidth]{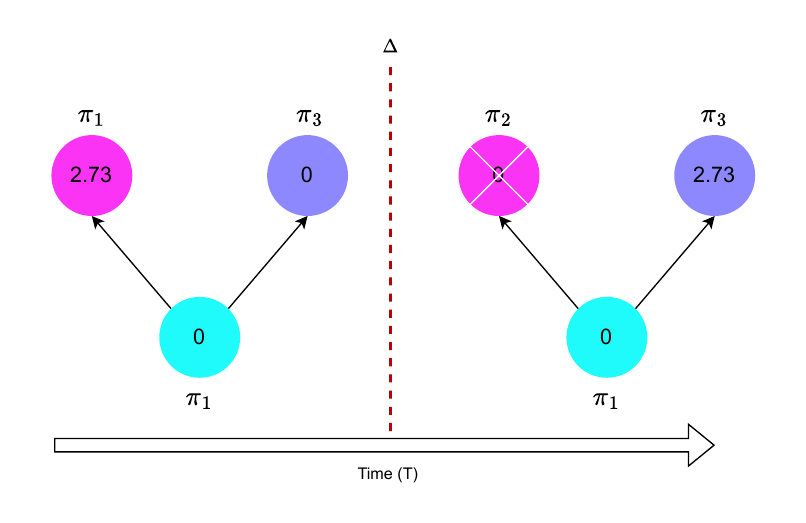}
    \caption{Experimental setup}
    \label{fig:enter-label}
\end{figure}
\label{environments}
We consider two types of environments as instances of the multi-armed bandit problem as described in preliminaries (see Sec. \ref{mab-cl}). 

The first environment represents a multi-armed stateless bandit setting. This environment returns rewards sampled from the hidden distribution $r(a)$ when $a$ is pulled, with Gaussian noise applied ($\alpha = 0.1$).
The second environment is a $2$-state multi-armed bandit setting. This environment also returns rewards sampled from the hidden distribution $r(a)$ when $a$ is pulled, with Gaussian noise applied ($\alpha = 0.1$). When epoch $T=\Delta$ is reached, the environment changes state, therefore returning the rewards sampled from $r_1(a)$, which is a permutation of $r(a)$.
In both environments, $a_1$ is selected as the starting arm, with $r(1)= 0$ and $r_1(1)=0$. Therefore, the initial distribution is $\pi_1 = 0.9$ and $\pi_i = \frac{0.1}{K-1}\quad \text{for } i = 2,\dots,K.$

\subsection{Model Validation in Static Environment}
\label{validation}
To validate our model, we compare its predictions with empirical data gathered from controlled experiments in \cite{madirolas2023caenorhabditis}. In these experiments, four food patches were arranged in a square at $1.2$ cm from the center of a plate. Each patch contained E. coli $OP50$ at distinct densities: $0.2$, $0.1$, $0.05$, and $0.025$ (measured as Optical Density). In each trial, worms were released from the center of the plate, and their distribution was recorded over $7200$ seconds. We compared the results of the experiment with the prediction of our model in the multi-armed stateless bandit setting.
\begin{figure}
    \centering
    \includegraphics[width=0.9\linewidth]{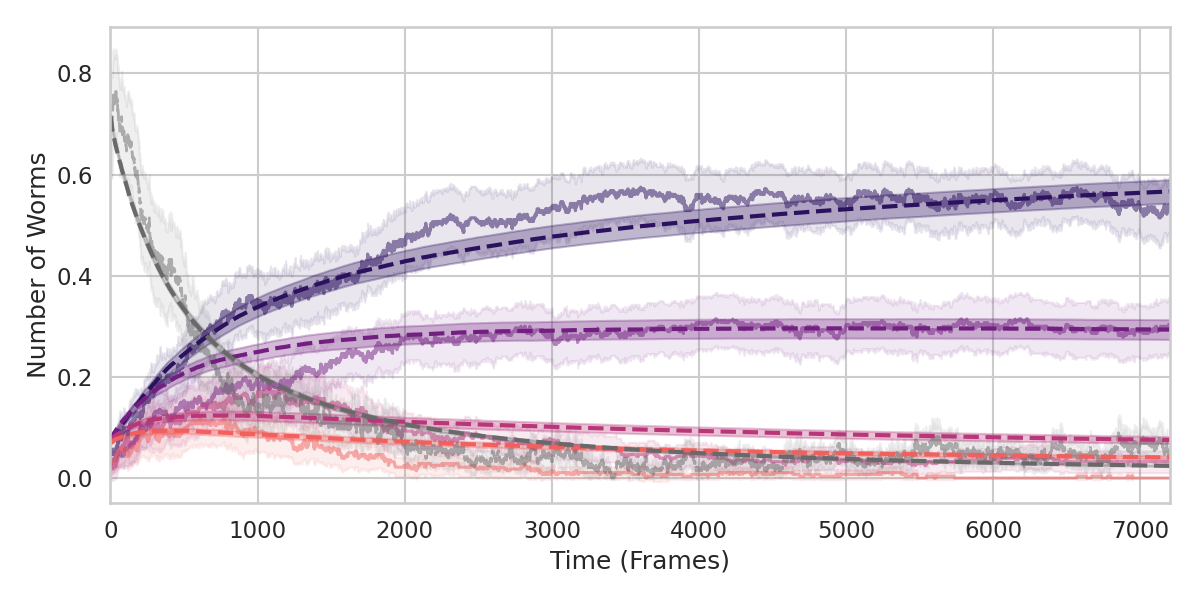}
    \caption{Proportion of worms in each patch (or outside the patches) as a function of time for two experiments with 4 food patches. Averaged over 58 videos, with 299 worms in total. Shaded areas show the 95\% confidence intervals, computed via bootstrapping. Dashed line represent the model predictions.}
    \label{fig:matching}
\end{figure}

Figure~\ref{fig:matching} displays the average time evolution of the proportion of worms present in each patch (as well as those outside the patches). The experimental curves illustrate how worms progressively distribute in patches, leading to an Ideal Free Distribution (IDF) \cite{HOUSTON1987301} .

We found the best-fit parameters using Differential Evolution (DE), with the search space bounded by the confidence intervals of parameters obtained in experiments for the attractiveness sigmoid function. The mean squared error ($MSE=\sum_{i=1}(x_i-y_i)^2$) was used as the fitness function. The parameters found for the Attractiveness sigmoid function are: $H=51.5, k=0.29, D_{attract}=0.003$.
The pheromone secretion parameter (i.e., the learning rate) of the model was also adjusted to best fit the data, therefore $Q$ was set to $0.02$. With these parameters, our model achieved an MSE of $2.595\mathrm{E}{-2}$, demonstrating close alignment between empirical data (shaded) and model predictions (dashed) and validating our model’s viability.

\subsection{Adaptation Dynamics in a Dynamic Environment}

To investigate the swarm’s adaptability under changing conditions, we use the $2$-state multi-armed bandit environment presented in \ref{environments} with three arms and the parameters found in \ref{validation}. 

Initially, arm 1 ($a_1$) has an attractiveness of $r(1) = 0$, $a_2$ has an attractiveness of $r(2) = 2.73$ (i.e., OP50 $OD=1$), and $a_3$ is set to $r(3) =0$. At epoch~$T=\Delta$, the environment transitions to a second state. In this new state, $a_1$ remains at attractiveness~$r_1(1) = 0$, $a_2$ resets to attractiveness~$r_1(2)=0$, and $a_3$ is updated to attractiveness~$r_1(3) = 2.73$.

\begin{figure}[!ht]
    \centering
    \includegraphics[width=1\linewidth]{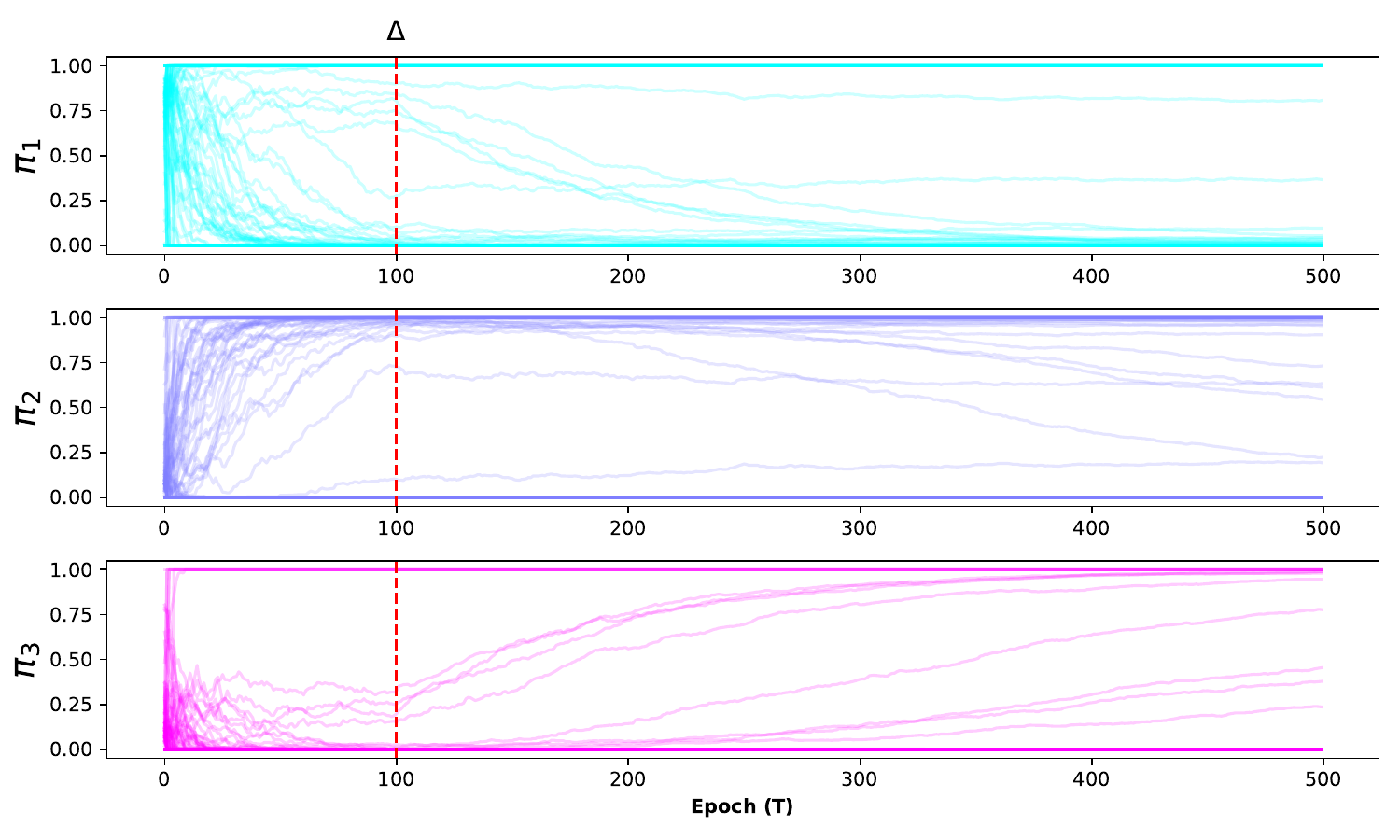}
    \caption{Policy evolution across 500 epochs for a homogeneous population. The swarm begins at $a_1$ with $\pi_1 \approx0.9$, moves to $a_2$ as the first swarming spot, and $a_3$ quality changes at the environmental switch (red line) at $\Delta=100$. Each line shows a single experiment, with $N=100$.}
    \label{fig:policy_homogeneous}
\end{figure}

Using this setup with $\Delta=100$, \textit{Memory size}$=350$, maximum number of epochs $T=500$, and number of runs $N=100$, we first examined a homogeneous population. As shown in Figure~\ref{fig:policy_homogeneous}, only $13\%$ of the runs successfully aggregated at the new spot after the state change. This limited success arises from the swarm’s strong positive stigmergic reinforcement: the continuous deposition of pheromones sustains a positive feedback loop that can trap the swarm at the original spot, even after its attractiveness has dropped to zero. 
A completely greedy policy prevents the agent to learn whether little-tried and so-far low value actions might lead to higher rewards. That is, it does not allow extinction of learned states. If the swarm is to acquire an accurate value function, greedy policies need to be modified to allow exploratory actions, at least occasionally. Greedy policies thus modified are called $\epsilon$-greedy \cite{sutton1998reinforcement}. 
Therefore, inspired by the $\epsilon$-greedy strategy, we introduce exploratory individuals into the population.

\begin{figure}[!ht]
    \centering
    \includegraphics[width=1\linewidth]{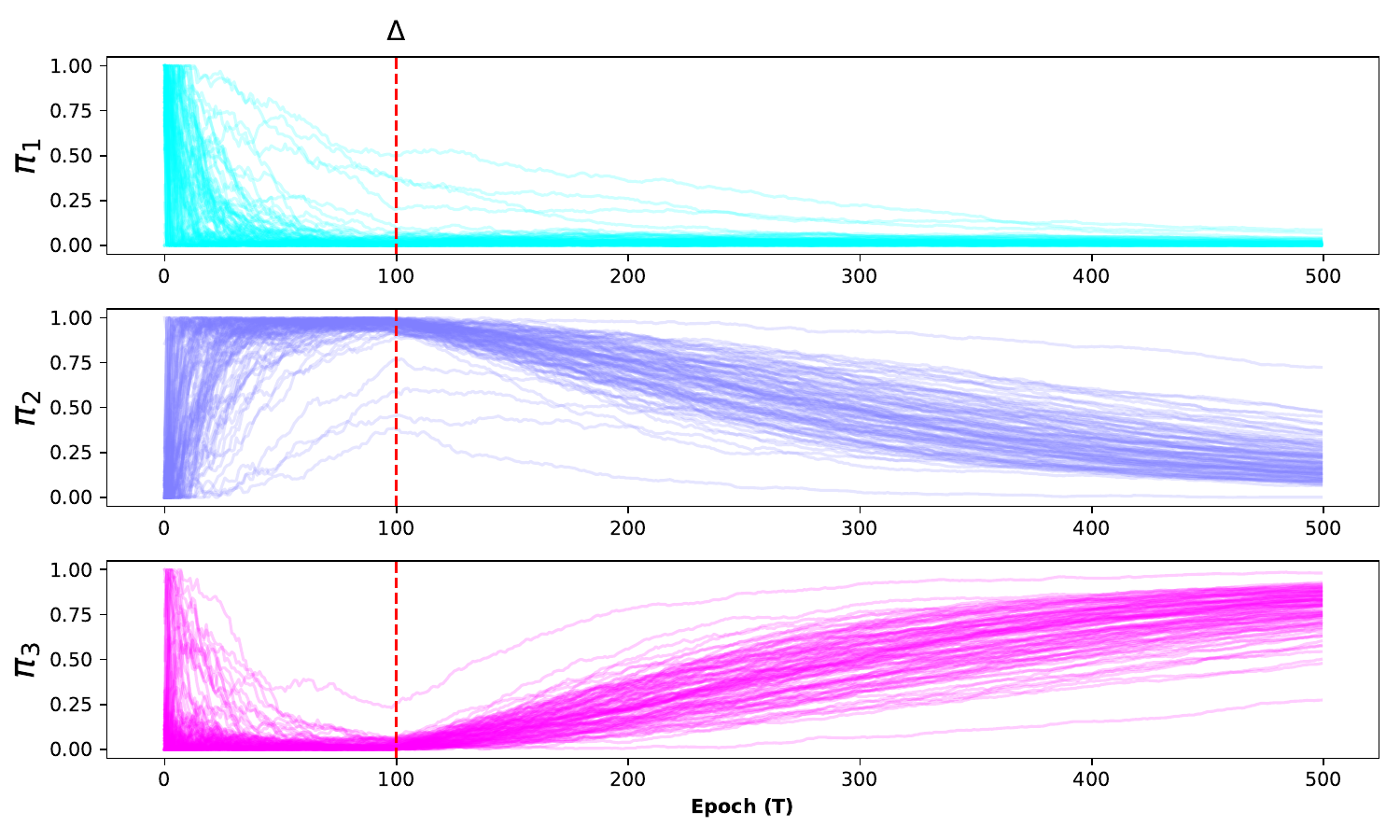}
    \caption{Policy evolution across 500 epochs for a heterogeneous population with $\epsilon=0.1$ of the individuals exploring. $a_1$ is the starting spot, $a_2$ is the first swarming spot, and $a_3$ quality changes at the environmental switch (red line) at $\Delta=100$. Each line shows one experiment, with $N=100$.}
    \label{fig:policy_heterogeneous}
\end{figure}

In the heterogeneous population, we assume that $1-\epsilon$ of the \celeg~can both sense and secrete pheromones, while $\epsilon$ are blind to pheromones but still secrete them. Hence, the exploratory group relies solely on bacterial density for navigation. Figure~\ref{fig:policy_heterogeneous} shows that introducing $\epsilon=0.1$ exploratory individuals leads to a 100\% success rate in reaching and aggregating at the final spot. These findings highlight the importance of exploratory agents in avoiding local attractor traps created by purely attractive pheromone-driven learning.

We now examine the parameters that influence the swarm’s adaptive capability.
To quantify the swarm’s response to the state change, we define the Mean Time to Adapt (MTA):
\begin{equation}
\mathrm{MTA} = \frac{1}{N} \sum_{i=1}^{N} 
\begin{cases}
k, & \text{if } \pi_3(\Delta + k) \ge 0.9,\\
T, & \text{otherwise},
\end{cases}
\end{equation}
where $\Delta$ is the epoch at which the environment changes, $N$ is the total number of runs, $T$ is the maximum number of epochs (i.e., the terminal time for the experiment), and $k$ is the first time (counted from $\Delta$) that $\pi_3$ reaches at least $0.9$, i.e., a $90\%$ policy concentration, indicating consensus in the swarm. If this threshold is never met, then $k$ is set to $T$.

\begin{figure}
    \centering
    \includegraphics[width=1\linewidth]{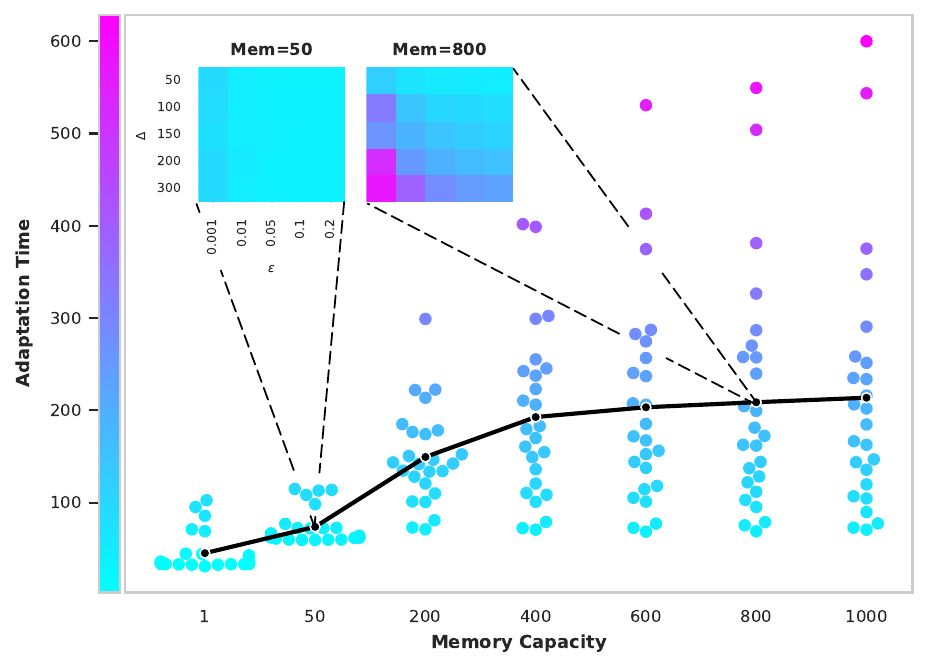}
    \caption{Adaptation to environment changes under varying memory capacity, environmental time swapping and population heterogeneity.
    We simulated $1000$ epochs of foraging under a dynamic two-patch environment, switching patch quality at different $\Delta$ values. Results were averaged over $5$ runs for each parameter set. 
    The swarmplot represents the Global mean adaptation time (MTA) as a function of pheromone persistence (\textit{memory capacity}). Average is shown by the black line. Heatmaps depict the proportion of heterogeneity ($\epsilon$) against the time spent before environment change ($\Delta$), colors represent the MTA. The x-axis denotes the fraction of exploratory individuals $\epsilon \in \{0.001,0.01,0.05,0.1,0.2\}$, and the y-axis shows the time spent in the initial environment before switching ($\in \{50,100,150,200,300\}$ epochs). 
    }
    \label{fig:global}
\end{figure}

Figure~\ref{fig:global} shows how the swarm’s mean time to adapt depends on the pheromone memory capacity, the epoch at which the state change occurs ($\Delta$), and the fraction of exploratory individuals ($\epsilon$). Two main regimes emerge from these experiments. 

When the memory capacity is low, pheromones evaporate quickly. Because they do not accumulate for long, the swarm remains flexible and easily relocates after the environment changes. In this volatile memory regime, Figure~\ref{fig:global} suggests that the time to adapt remains nearly constant and does not depend strongly on $\Delta$ or $\epsilon$. This is because older pheromones vanish before they can trap the swarm in a previously rewarding spot.
Gradually, until the memory capacity reaches $\approx400$, pheromone persistence plays a diminishing role in the swarm’s adaptive capability.
When the memory capacity exceeds $400$, we can observe a plateau, where pheromones persist in the environment for much longer. This persistent memory means that the swarm is more likely to become locked onto locations it reinforced early on, even if that spot is no longer attractive. In this regime, both $\Delta$ and $\epsilon$ matter significantly. Shorter periods in the initial environment and a greater fraction of exploratory individuals reduce the chance that the swarm stays on an outdated patch. As seen in the heatmap where Memory capacity $=800$, even introducing a small set of non-pheromone-guided individuals (i.e., $0.01$) can help the swarm overcome strong pheromone signals and transition to the newly rewarding location. Heatmaps for Memory capacities greater than $400$ are duplicates of the one presented in Fig.\ref{fig:global} for Memory capacity  $=800$, and are therefore not shown.

\section{Conclusion}
The formalism presented in this work offers a new lens for interpreting stigmergic behaviour as a form of reinforcement learning. By establishing an equivalence between pheromone-modulated aggregation and cross-learning updates, our model suggests that future engineered \celeg\ populations (and by extension, other stigmergic systems) could implement a primitive form of collective and distributed operant conditioning. This equivalence, however, rests on strong abstractions: for instance, we assume linear, unbounded attraction to pheromone concentration and fixed bacterial patch attractiveness, neglecting sensory saturation, adaptation, or aversive responses known in real nematode behaviour. These simplifications enabled analytical tractability, but they limit the model’s biological realism. Incorporating nonlinear sensory responses or time-dependent changes in valence (e.g., switching from attraction to repulsion) remains an important avenue for future work.

By deriving an update rule structurally identical to classical cross-learning, we show that positive stigmergy (pheromone accumulation toward a site) acts as an implicit reward signal, converting collective aggregation into an operant-like learning process. This reconceptualises pheromone trails not merely as coordination tools but as externalised memory structures enabling distributed credit assignment and incremental policy revision. Such abstraction, while simplifying biological realism, confirms a general principle: collective decision-making can emerge from repeated, reward-modulated interactions between agents and their environment, with the environment itself storing reinforcement history, principles reminiscent of Ant Colony Optimization (ACO) \cite{dorigo2018ant} and other stigmergic systems.

We have presented a unified theoretical framework that embeds stigmergic coordination within the broader paradigm of reinforcement learning. By demonstrating the mathematical equivalence between pheromone-driven site selection and cross-learning updates, we show that engineered \celeg\ swarms may, in principle, embody collective cognitive learning dynamics through environmentally mediated reinforcement.

Moreover, we reveal that while stigmergy enables efficient convergence in static environments, it can hinder adaptability in dynamic settings unless mechanisms for extinction and exploration are integrated. Behavioural heterogeneity, operationalised through a minority of exploratory individuals, restores flexibility and allows the swarm to revise outdated collective decisions.

Our results not only unveil theoretical links between biological learning, swarm dynamics, and RL, but also offer actionable insights for the design of biohybrid systems and decentralised algorithms. Future efforts will focus on implementing task-switching strategies within homogeneous populations, comparing these results with other heterogenous strategies and validating these principles in living \celeg\ systems and in agent-based simulations.

\section{Acknowledgements}
The BABots project has received funding from the Horizon Europe, PathFinder European Innovation Council Work Programme under grant agreement No 101098722. Views and opinions expressed are however those of the authors only and do not necessarily reflect those of the European Union or European Innovation Council and SMEs Executive Agency (EISMEA). Neither the European Union nor the granting authority can be held responsible for them.

\appendix
\section{Cross-learning equivalence derivation}
\subsection{Case 1: Site i is Chosen}
\label{deriv1}
Assume site $i$ is chosen at time $t$. Then
\begin{equation}
  \tau_i(t+1) = \rho\,\tau_i(t) + Q, 
  \qquad
  \tau_k(t+1) = \rho\,\tau_k(t)
  \quad (\forall\,k \neq i).
\end{equation}
Only $\tau_i(t)$ receives an increment of $ Q $, while all other pheromone levels $\tau_k(t)$ for $k \neq i$ are solely subject to evaporation by the factor $ \rho $. Therefore, when normalizing to compute $ P_i(t+1) $, the additional pheromone $ Q $ contributes only to the chosen site's term, resulting in $ Q A_i $ in the denominator.
Hence the probability of choosing site $i$ at time $t+1$ becomes:
\begin{equation}
\label{eq:Pi-t+1-chosen-rev}
  P_i(t+1)
  \;=\;
  \frac{\bigl[\rho\,\tau_i(t) + Q\bigr]\,A_i}
       {\rho\,\sum_{j=1}^M \tau_j(t)\,A_j 
        \;+\; 
        Q\,A_i}.
\end{equation}
We rewrite $\tau_i(t)$ in terms of $P_i(t)$ by noting
\begin{equation}
      \tau_i(t) 
  \;=\; 
  \frac{P_i(t)}{A_i} 
  \,\Bigl[\sum_{j=1}^M \tau_j(t)\,A_j\Bigr].
\end{equation}

Substitute into \eqref{eq:Pi-t+1-chosen-rev}:
\begin{equation}
\begin{aligned}
  P_i(t+1)
  &=\; 
  \frac{\rho\,\tau_i(t)\,A_i + Q\,A_i}
       {\rho\,\sum_{j=1}^M \tau_j(t)\,A_j + Q\,A_i} 
  \\[6pt]
  &=\; 
  \frac{\rho\,\bigl[\tfrac{P_i(t)}{A_i}\sum_{j=1}^M \tau_j(t)\,A_j\bigr]A_i + Q\,A_i}
       {\rho\,\sum_{j=1}^M \tau_j(t)\,A_j + Q\,A_i} 
  \\[6pt]
  &=\;
  \frac{\rho\,P_i(t)\,\sum_{j=1}^M \tau_j(t)\,A_j + Q\,A_i}
       {\rho\,\sum_{j=1}^M \tau_j(t)\,A_j + Q\,A_i}.
\end{aligned}
\end{equation}
Define 
\begin{equation}
  r 
  \;=\;
  \frac{Q\,A_i}{\rho\,\sum_{j=1}^M \tau_j(t)\,A_j + Q\,A_i}.
\end{equation}
Rewriting to highlight the increment in $P_i(t)$ yields
\begin{equation}
  P_i(t+1)
  \;=\; 
  P_i(t)
  \;+\;
  r
  \,[\,1 - P_i(t)].
\end{equation}
Thus, if site $i$ is chosen, $P_i(t)$ is incremented by a factor proportional to $[1 - P_i(t)]$.

\subsection{Case 2: Site i is Not Chosen}
\label{deriv2}
If site $i$ is not chosen at time $t$, then there is no new pheromone added to $\tau_i(t)$, so
\begin{equation}
  \tau_i(t+1) \;=\; \rho\,\tau_i(t).
\end{equation}
Meanwhile, exactly one other site (say $k \neq i$) is chosen and thus gets reinforced.  Hence, 
\begin{equation}
\begin{aligned}
    \tau_k(t+1) 
  = 
  \rho\,\tau_k(t) \;+\; Q,
  \\
  \text{and for all others } j\neq \{i,k\}:\\
  \tau_j(t+1)=\rho\,\tau_j(t).  
\end{aligned}
\end{equation}
We can now convert these $\tau_i(t+1)$ values to updated probabilities by normalizing over the sum of all pheromone$\times$attractiveness terms:
\begin{equation}
  P_i(t+1) 
  \;=\; 
  \frac{\rho\,\tau_i(t)\,A_i}
       {\displaystyle \sum_{j=1}^M \bigl[\rho\,\tau_j(t) + \Delta_j\bigr]\,A_j},
\end{equation}
where 
$\Delta_j = Q$ if $j$ was chosen, and $0$ otherwise.  
Since exactly one site $k$ is chosen,
\begin{equation}
  \sum_{j=1}^M \bigl[\rho\,\tau_j(t) + \Delta_j\bigr] A_j
  \;=\;
  \rho\,\sum_{j=1}^M \tau_j(t)\,A_j 
  \;+\; 
  Q\,A_k.
\end{equation}
Hence
\begin{equation}
  P_i(t+1)
  \;=\;
  \frac{\rho\,\tau_i(t)\,A_i}
       {\rho\,\sum_{j=1}^M \tau_j(t)\,A_j + Q\,A_k}.
\end{equation}

Expressing $\tau_i(t)$ in terms of $P_i(t)$:

\begin{equation}
\begin{aligned}
  P_i(t)
  &=
  \frac{\tau_i(t)\,A_i}{\sum_{j=1}^M \tau_j(t)\,A_j}\\
  &\quad\Longrightarrow\quad
  \tau_i(t) 
  \;=\; 
  \frac{P_i(t)}{A_i}\,\Bigl[\sum_{j=1}^M \tau_j(t)\,A_j\Bigr].
  \end{aligned}
\end{equation}
Substitute back to get
\begin{equation}
\begin{aligned}
  P_i(t+1)
  &=
  \frac{\rho\,\bigl[\tfrac{P_i(t)}{A_i}\sum_{j=1}^M \tau_j(t)\,A_j\bigr]\,A_i}
       {\rho\,\sum_{j=1}^M \tau_j(t)\,A_j + Q\,A_k}\\
  &= P_i(t)\,
  \frac{\rho\,\sum_{j=1}^M \tau_j(t)\,A_j}
       {\rho\,\sum_{j=1}^M \tau_j(t)\,A_j + Q\,A_k}.
\end{aligned}
\end{equation}
Factor out $P_i(t)$ and rewrite the fraction:
\begin{equation}
  P_i(t+1)
  \;=\;
  P_i(t)\,\biggl( 1 \;-\; 
  \frac{Q\,A_k}{\,\rho\,\sum_{j=1}^M \tau_j(t)\,A_j + Q\,A_k\,}
  \biggr).
\end{equation}
Define 
\begin{equation}
  r 
  \;=\;
  \frac{Q\,A_k}{\rho\,\sum_{j=1}^M \tau_j(t)\,A_j + Q\,A_k}.
\end{equation}
Then the update law becomes
\begin{equation}
  P_i(t+1)
  \;=\;
  P_i(t)
  \;-\;
  r\,P_i(t),
\end{equation}
showing that if site $i$ is not chosen, its probability $P_i(t)$ decreases by a fraction $r$ proportional to $P_i(t)$ itself.


\end{document}